\pdfoutput=1
\documentclass[10pt,twocolumn,letterpaper]{article}

\usepackage[utf8]{inputenc}
\usepackage[T1]{fontenc}
\usepackage{charter}
\usepackage{amsmath,amssymb,amsfonts}
\usepackage{mathtools}
\usepackage{microtype}

\usepackage[
  letterpaper,
  top=2.0cm,
  bottom=1.9cm,
  left=1.75cm,
  right=1.75cm,
  headheight=15pt,
  headsep=12pt,
  footskip=22pt
]{geometry}
\usepackage{xcolor}
\definecolor{pcText}{HTML}{111111}
\definecolor{pcSubtle}{HTML}{222222}
\definecolor{pcMuted}{HTML}{333333}
\definecolor{pcRule}{HTML}{222222}
\definecolor{pcLink}{HTML}{174EA6}
\definecolor{pcBoxBg}{HTML}{F4F4F4}
\definecolor{pcCodeBg}{HTML}{F8F9FA}

\AtBeginDocument{%
  \color{pcText}%
  \fontsize{9.5pt}{11.6pt}\selectfont%
}

\usepackage{cite}
\usepackage{url}
\usepackage[
  colorlinks=true,
  linkcolor=pcLink,
  citecolor=pcLink,
  urlcolor=pcLink,
  pdfauthor={Wesley Kambale and Roland Sankara},
  pdftitle={Pearl-Chat: Engineering a native Luganda language model from first principles in JAX and Flax NNX}
]{hyperref}

\usepackage{graphicx}
\usepackage{booktabs}
\usepackage{tabularx}
\usepackage{array}
\usepackage{float}
\usepackage{dblfloatfix}

\newcolumntype{L}[1]{>{\hsize=#1\hsize\raggedright\arraybackslash}X}
\newcommand{\pctablefont}{\fontsize{8.5pt}{10.4pt}\selectfont}

\usepackage{caption}
\DeclareCaptionLabelSeparator{pipe}{ \textbar\ }
\newcounter{algorithm}
\newcommand{\algorithmbox}[4]{%
  \par\vspace{0.50em}%
  \noindent\begin{minipage}{\columnwidth}%
    \refstepcounter{algorithm}%
    \vspace{2pt}%
    \noindent\rule{\columnwidth}{1.0pt}\par\vspace{3pt}%
    {\fontsize{8.2pt}{10.2pt}\selectfont%
      \noindent\textbf{#1}\par\vspace{2pt}%
      \noindent\rule{\columnwidth}{0.5pt}\par\vspace{3pt}%
      \noindent\textbf{Input:} #2\par%
      \noindent\textbf{Output:} #3\par\vspace{2pt}%
      \noindent{\color{pcMuted}\rule{\columnwidth}{0.4pt}}\par\vspace{3pt}%
      #4%
    }%
    \par\vspace{2pt}%
    \noindent\rule{\columnwidth}{1.0pt}\par\vspace{3pt}%
  \end{minipage}%
}
\newcommand{\algline}[2]{%
  \noindent\makebox[1.4em][r]{{\fontsize{7.6pt}{9.2pt}\selectfont\color{pcMuted}#1:}}\hspace{0.4em}#2\par%
}
\newcommand{\algcaption}[1]{%
  \par\vspace{2pt}%
  \noindent{\fontsize{8.6pt}{10.5pt}\selectfont\textbf{Algorithm \thealgorithm} \textbar\ #1}\par\vspace{0.55em}%
}

\usepackage{titlesec}
\renewcommand{\thesection}{\arabic{section}.}
\renewcommand{\thesubsection}{\arabic{section}.\arabic{subsection}.}
\renewcommand{\thesubsubsection}{\arabic{section}.\arabic{subsection}.\arabic{subsubsection}.}

\titleformat{\section}
  {\fontsize{11.2pt}{13.5pt}\selectfont\bfseries\color{black}}
  {\thesection}{0.45em}{}
\titlespacing*{\section}{0pt}{0.92em}{0.45em}

\titleformat{\subsection}
  {\fontsize{9.8pt}{12.0pt}\selectfont\bfseries\color{black}}
  {\thesubsection}{0.45em}{}
\titlespacing*{\subsection}{0pt}{0.76em}{0.36em}

\titleformat{\subsubsection}
  {\fontsize{9.5pt}{11.5pt}\selectfont\bfseries\itshape\color{black}}
  {\thesubsubsection}{0.45em}{}
\titlespacing*{\subsubsection}{0pt}{0.65em}{0.32em}

\usepackage{enumitem}
\setlist[itemize]{itemsep=0.20em, topsep=0.32em, parsep=0pt, leftmargin=1.4em}
\setlist[enumerate]{itemsep=0.20em, topsep=0.32em, parsep=0pt, leftmargin=1.5em}

\newenvironment{pcquote}{%
  \par\vspace{0.38em}%
  \begin{quote}\setlength{\parindent}{0pt}\setlength{\leftmargin}{1.0em}\setlength{\rightmargin}{0.5em}%
}{%
  \end{quote}%
  \par\vspace{0.38em}%
}

\usepackage{listings}
\usepackage{fancyhdr}
\fancypagestyle{firstpage}{%
  \fancyhf{}

  \fancyfoot[L]{%
    {\fontsize{7.8pt}{9.8pt}\selectfont\color{pcSubtle}%
      \textsuperscript{*}Corresponding author: \href{mailto:kambale@aiforuganda.org}{kambale@aiforuganda.org}\\
      \copyright\ 2026 AI for Uganda Initiative. All rights reserved.%
    }%
  }
}

\newcommand{\makepctitle}[1]{%
  \noindent\begin{minipage}{\textwidth}%
    \setlength{\parindent}{0pt}%
    \noindent\begin{tabular*}{\textwidth}{@{\extracolsep{\fill}}lr@{}}
      \includegraphics[height=0.92cm]{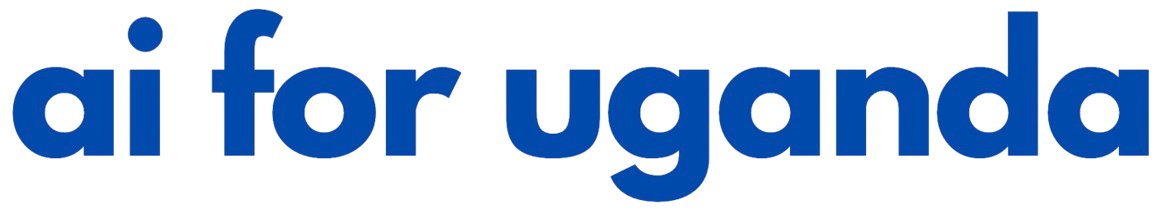} &
      {\fontsize{8.5pt}{10pt}\selectfont\color{pcText}2026-10-12}
    \end{tabular*}\par\vspace{4pt}%
    \noindent\rule{\textwidth}{0.6pt}\par\vspace{0.75em}%
    {\fontsize{17.2pt}{21pt}\selectfont\bfseries\color{black}%
      Pearl-Chat: Engineering a native Luganda language model from first principles in JAX and Flax NNX\par}%
    \vspace{0.65em}%
    {\fontsize{10.2pt}{13pt}\selectfont\bfseries\color{pcText}%
      Wesley Kambale\textsuperscript{1,*} and Roland Sankara\textsuperscript{2}\par}%
    \vspace{0.18em}%
    {\fontsize{9.0pt}{11.5pt}\selectfont\color{pcSubtle}%
      \textsuperscript{1}AI for Uganda Initiative, \textsuperscript{2}ISBAT University\par}%
    \vspace{0.80em}%
    {\fontsize{9.2pt}{12.5pt}\selectfont\bfseries\color{pcText}%
      #1\par}%
    \vspace{0.40em}%
    {\fontsize{8.5pt}{11pt}\selectfont\color{pcMuted}%
      \textbf{Keywords:} Natural language processing, low-resource languages, Bantu morphology, Luganda, JAX, Flax NNX, byte-level BPE, functional neural state, two-stage curriculum, open science.\par}%
    \vspace{1.15em}%
  \end{minipage}%
}

\newcommand{\contributionsbox}[1]{%
  \par\vspace{0.45em}%
  \noindent\fcolorbox{black}{pcBoxBg}{%
    \begin{minipage}{\dimexpr\columnwidth-18pt-1.8pt\relax}%
      \vspace{4pt}%
      \setlength{\parindent}{0pt}%
      {\fontsize{8.7pt}{11.2pt}\selectfont #1}%
      \vspace{4pt}%
    \end{minipage}%
  }\par\vspace{0.45em}%
}

\begin{document}

\twocolumn[{%
\makepctitle{%
Autoregressive foundation language models predominantly target high-resource Indo-European languages, imposing severe computational and morphological penalties on under-represented African languages. Generic multilingual tokenizers routinely suffer from high fertility rates exceeding 3.2 tokens per word on agglutinative Bantu languages, degrading context efficiency and inflating inference latency. This paper presents Pearl-Chat, an 8,473,122-parameter ($\sim$8.5M) causal decoder-only transformer engineered specifically for Luganda (Oluganda, ISO 639-2: lg), a low-resource Niger-Congo Bantu language spoken by over 10 million people in East Africa. Built from first mathematical principles in JAX and Flax NNX, Pearl-Chat avoids monolithic black-box abstractions, demonstrating explicit functional state decomposition (\texttt{nnx.split} and \texttt{nnx.merge}), deterministic resumable data streaming through Google Grain, and SPMD tensor parallelism using \texttt{jax.sharding.Mesh}. We introduce a custom byte-level byte-pair encoding (BPE) subword tokenizer trained from scratch on curated Luganda corpora ($V = 8,192$), achieving a compression ratio of 4.88 characters per token and a sequence fertility of 1.60 tokens per word. This represents a two-fold compression improvement over LLaMA-3 (128K) and Gemma (256K). The model is trained using a two-stage curriculum: Stage 1 foundational pre-training on 59,501 sequence records across parallel, biblical, encyclopedic, and journalistic corpora, followed by Stage 2 conversational instruction fine-tuning on 510 dialogues with assistant-only loss masking. During training across 4,000 optimization steps, cross-entropy validation loss drops monotonically from an uninitialized baseline of 9.4695 (perplexity: 12,958.98) to 0.1818 (perplexity: 1.20), yielding an absolute decrease of 9.2877 nats. In a three-tier held-out diagnostic benchmark, Pearl-Chat achieves a held-out ROUGE-L of 0.9792 and BLEU of 0.9405 while demonstrating grammatical concordial agreement across Bantu noun classes. We provide a lightweight Python Server-Sent Events (SSE) streaming runtime and open-source all weights, tokenizers, and interactive environments to establish an open engineering standard for African language model research.%
}%
}]
\thispagestyle{firstpage}

\section{Introduction}
The global landscape of natural language processing exhibits a severe digital divide \cite{nekoto2020participatory}. Despite the existence of over 7,000 living spoken languages worldwide, more than 95\% of contemporary large language model compute, pre-training token volume, and architectural optimization focus on a small subset of high-resource Indo-European languages \cite{brown2020language, touvron2023llama, dubey2024llama3}. The African continent, representing over 2,000 distinct languages and 1.4 billion people, remains under-represented in modern foundation model architectures \cite{orife2020masakhane, adelani2022masakhaner}.

Among East African languages, Luganda (Oluganda, ISO 639-2: lg) is a major Niger-Congo language of the Narrow Bantu branch (JE.15), spoken by more than 10 million people across Uganda and the northern Lake Victoria basin \cite{ashton1954luganda, hyman2003bantu}. Luganda exhibits rich agglutinative verbal morphology, an extensive noun-class system comprising 23 distinct nominal genders with alliterative concordial agreement, phonemic vowel length, and consonant gemination. When state-of-the-art multilingual language models such as LLaMA-3 \cite{dubey2024llama3}, GPT-4 \cite{openai2023gpt4}, or Gemma \cite{gemmateam2024gemma} process Luganda text, they encounter what is recognized in multilingual and African NLP research as the tokenizer tax \cite{ahia2023all, adelani2022masakhaner}. Because their massive vocabularies ($V \in [128\text{K}, 256\text{K}]$) are trained predominantly on English, European, and East Asian text, Luganda words are fragmented into isolated characters or arbitrary byte sequences. This fragmentation inflates token fertility to between 3.2 and 3.9 tokens per word, truncating effective context length, inflating inference costs by up to 400\%, and degrading semantic representations \cite{sennrich2016neural, ahia2023all}.

Simultaneously, machine learning engineering has become heavily reliant on monolithic, stateful frameworks. PyTorch abstractions frequently rely on implicit object mutation, hidden runtime hooks, and imperative graph interception that obscure the underlying mathematical formulation of deep learning \cite{bradbury2018jax, heek2024flax}. In contrast, the JAX ecosystem formulates deep learning as composable mathematical function transformations:
\begin{equation*}
f: (\theta, x) \mapsto (\hat{y}, \theta')
\end{equation*}
leveraging Accelerated Linear Algebra (XLA) compilation, functional gradient transformations (\texttt{jax.grad}), automatic vectorization (\texttt{jax.vmap}), and declarative Single Program Multiple Data (SPMD) distributed tensor sharding (\texttt{jax.sharding.Mesh}) \cite{bradbury2018jax}.

In this work, we present Pearl-Chat, an autoregressive decoder-only language model engineered from first mathematical principles specifically for Luganda, developed as part of the advanced research track at PyCon Africa 2026 \cite{kambale2026pearlchat}. Pearl-Chat is implemented in JAX and Flax NNX, eliminating third-party modeling abstractions and PyTorch dependencies.

\contributionsbox{%
The core contributions of this work are summarized as follows:
\begin{enumerate}
  \item Linguistically aligned native tokenizer: We train a custom byte-level BPE tokenizer ($V = 8,192$) from scratch on native Luganda text, achieving a compression ratio of 4.88 characters per token and a sequence fertility of 1.60 tokens per word. This reduces sequence lengths by more than half compared to frontier multilingual models.
  \item JAX and Flax NNX architecture: We implement an 8.5M-parameter GPT-2 style transformer incorporating Pre-Layer Normalization, GELU activations, and parameter-efficient embedding weight tying, explicitly managing functional state through \texttt{nnx.split} and \texttt{nnx.merge}.
  \item Deterministic resumable streaming: We integrate Google Grain for preemption-resilient, deterministic data loading across four heterogeneous Luganda datasets totaling over 59,500 curated sequences.
  \item Two-stage optimization curriculum: We formulate a two-stage training curriculum combining Stage 1 foundational causal pre-training with Stage 2 assistant-masked conversational instruction tuning using Optax AdamW and cosine decay.
  \item Empirical verification and serving: We validate Pearl-Chat across a three-tier diagnostic benchmark (achieving a held-out ROUGE-L of 0.9792 and BLEU of 0.9405) and deploy a lightweight Server-Sent Events (SSE) streaming runtime published openly on GitHub and the Hugging Face Hub.
\end{enumerate}%
}

\section{Linguistic typology and morphological architecture}
Effective language model engineering for low-resource Bantu languages demands an appreciation of their underlying grammar. Luganda belongs to the Niger-Congo phylum, Narrow Bantu branch, sub-group JE.15 \cite{ashton1954luganda, hyman2003bantu}. Unlike isolating or fusional languages, Luganda is fundamentally agglutinative and polypersonal, encoding complex syntactic relations through affixes attached to verbal roots.

\subsection{Agglutinative verbal template}
In Luganda, a single orthographic word frequently functions as an entire multi-clause sentence. A finite verb follows a strict morphological template:
\begin{equation*}
\text{Verb} = \text{PreInit} - \text{SM} - \text{TAM} - \text{OM} - \text{Root} - \text{Ext} - \text{FV}
\end{equation*}
where PreInit represents initial negative or relative markers, SM is the subject marker concord, TAM represents tense, aspect, and mood markers, OM is the object marker infix, Root is the lexical verb stem, Ext represents derivational suffixes (causative, applicative, reciprocal, passive), and FV is the final vowel.

Consider the sentence from our benchmark corpus:
\begin{pcquote}
\emph{``Abayizi bagenda ku masomero okufuna amagezi n'obukugu obunaabayamba mu bulamu.''}
\end{pcquote}
In the word \emph{obunaabayamba} (``that will help them''):
\begin{enumerate}
  \item \emph{o-}: initial augment and pre-prefix (noun class 14)
  \item \emph{-bu-}: subject concord (class 14 abstract noun \emph{obukugu})
  \item \emph{-naa-}: future tense marker
  \item \emph{-ba-}: third person plural object concord (``them'', class 2 \emph{abayizi})
  \item \emph{-yamb-}: lexical root (``help'')
  \item \emph{-a}: final indicative vowel
\end{enumerate}

If a tokenizer fragments \emph{obunaabayamba} into arbitrary character n-grams, the transformer attention heads must spend substantial capacity reconstructing the basic morphological spine. Conversely, a native subword vocabulary identifies grammatical prefixes as unified tokens, preserving morphological compositionality.

\begin{table*}[t]
  \caption{Luganda noun class taxonomy and concordial agreement template.}
  \label{tab:noun_classes}
  \pctablefont
  \begin{tabularx}{\textwidth}{L{0.67} L{0.83} L{0.77} L{1.73}}
    \toprule
    \textbf{Class} & \textbf{Prefix} & \textbf{Concord} & \textbf{Example gloss} \\
    \midrule
    Cl 1 / 2 & omu- / aba- & a- / ba- & \emph{omuntu / abantu} (person / people) \\
    Cl 3 / 4 & omu- / emi- & gu- / gi- & \emph{omuti / emiti} (tree / trees) \\
    Cl 5 / 6 & eri- / ama- & li- / ga- & \emph{erinnya / amannya} (name / names) \\
    Cl 7 / 8 & eki- / ebi- & ki- / bi- & \emph{ekitabo / ebitabo} (book / books) \\
    Cl 9 / 10 & en- / ezin- & e- / zi- & \emph{ente / ente} (cow / cows) \\
    Cl 11 & olu- & lu- & \emph{olulimi} (language / tongue) \\
    Cl 12 / 13 & aka- / otu- & ka- / tu- & \emph{akambe} (small knife) \\
    Cl 14 & obu- & bu- & \emph{obulamu} (life / abstract) \\
    Cl 15 & oku- & ku- & \emph{okugenda} (to go / infinitive) \\
    Cl 16--18 & wa- / ku- / mu- & wa- / ku- / mu- & \emph{awaka, kungulu, munda} (locatives) \\
    \bottomrule
  \end{tabularx}
\end{table*}

\subsection{Concordial agreement and noun classes}
Luganda categorizes nouns into 23 distinct noun classes (\emph{ebibiina by'amannya}), grouped into singular and plural pairs (Table~\ref{tab:noun_classes}). Grammatical concord dictates that adjectives, numerals, possessives, demonstratives, and verbs must alliteratively agree with the head noun class:
\begin{pcquote}
\emph{Ekitabo kino ekirungi kigudde.} (Class 7 singular: ``This good book has fallen.'') \\
\emph{Ebitabo bino ebirungi bigudde.} (Class 8 plural: ``These good books have fallen.'')
\end{pcquote}
An autoregressive model must track these concord dependencies across multi-token spans. If the subject prefix shifts from \emph{ki-} to \emph{bi-}, the subsequent verb must reliably shift its agreement concord.

\subsection{Orthography, gemination, and tonal prosody}
Luganda orthography uses the Latin alphabet with phonemic distinctions:
\begin{enumerate}
  \item Geminate vowels: Double vowels (\emph{aa, ee, ii, oo, uu}) denote phonemic vowel length (\emph{okula} ``to grow'' versus \emph{okuula} ``to uproot'').
  \item Consonant gemination: Double consonants denote phonetic lengthening and tonal blockage, such as in \emph{nnyo} (``very''), \emph{ssebo} (``sir''), or \emph{ggye} (``army'').
  \item Elision and cliticization: Vowels elide across word boundaries, represented orthographically with apostrophes (for example, \emph{n'obukugu} from \emph{na + obukugu}).
\end{enumerate}

Generic tokenizers fail on these clitics, treating apostrophes as punctuation breaks rather than syntactic conjunctions.

\section{Byte-level BPE tokenizer engineering and compression analysis}
Tokenization governs the interface between continuous vector embeddings and symbolic linguistic structures \cite{sennrich2016neural, ahia2023all}.

\begin{figure*}[t]
  \centering
  \includegraphics[width=0.84\textwidth]{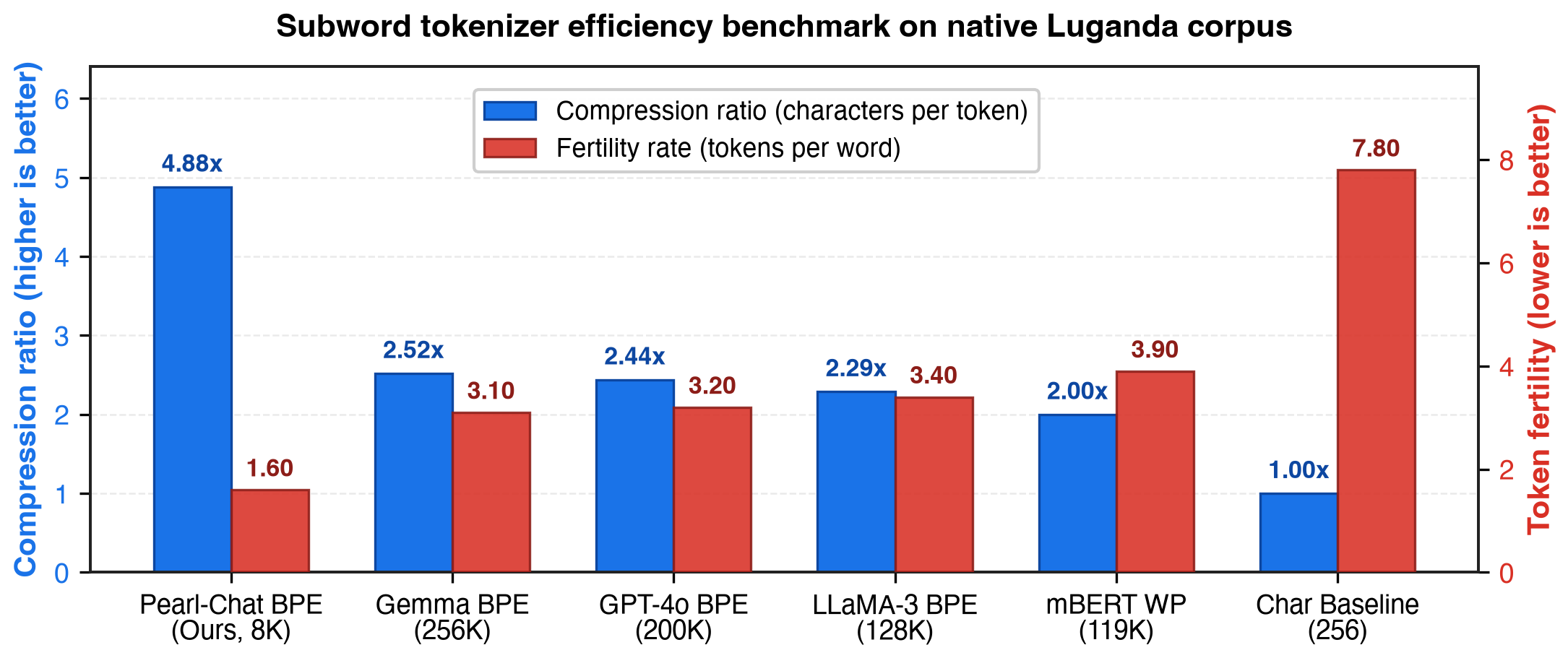}
  \caption{Subword tokenizer efficiency benchmark on native Luganda prose. Pearl-Chat BPE ($V=8,192$) achieves a 4.88 characters per token compression ratio and 1.60 tokens per word fertility, outperforming 100K+ multilingual vocabularies.}
  \label{fig:tokenizer_benchmark}
\end{figure*}

\subsection{Algorithmic formulation}
We train a custom byte-level byte-pair encoding (BPE) tokenizer from scratch using the Hugging Face \texttt{tokenizers} library (Algorithm~1). The base vocabulary starts with the 256 individual byte values of the UTF-8 encoding space:
\begin{equation*}
V_0 = \{ b_0, b_1, \dots, b_{255} \} \cup \{ \text{\texttt{<|endoftext|>}}, \text{\texttt{<|pad|>}}, \text{\texttt{<|unk|>}} \}
\end{equation*}
By grounding the vocabulary in raw bytes, out-of-vocabulary (OOV) tokens are prevented; arbitrary unicode strings are decomposed losslessly into base byte units.

The iterative merge operation calculates the pair frequency of adjacent symbols $(s_i, s_j)$ across the training corpus $C$:
\begin{equation*}
(s_a, s_b) = \operatorname*{arg\,max}_{(s_i, s_j)} \text{Freq}(s_i, s_j, C)
\end{equation*}
and introduces a new atomic token $s_{ab} = s_a \circ s_b$ into the vocabulary:
\begin{equation*}
V_{k+1} = V_k \cup \{ s_{ab} \}
\end{equation*}
Merge iterations proceed until the target vocabulary size $V = 8,192$ ($2^{13}$) is attained.

\algorithmbox{Algorithm 1: Byte-level BPE vocabulary construction and morphological merge loop}%
  {Raw Luganda corpus $\mathcal{C}$, Target vocabulary capacity $V = 8,192$, Special tokens $\mathcal{S}$}%
  {Trained subword vocabulary $\mathcal{V}$ and ordered merge table $\mathcal{M}$}%
  {%
    \algline{1}{\textbf{procedure} \textsc{TrainByteLevelBPE}$(\mathcal{C}, V, \mathcal{S})$}
    \algline{2}{\quad $\mathcal{V}_0 \leftarrow \{ b_0, b_1, \dots, b_{255} \} \cup \mathcal{S}$}
    \algline{3}{\quad $\mathcal{M} \leftarrow []$}
    \algline{4}{\quad $\mathcal{D} \leftarrow \text{PreTokenizeUTF8Bytes}(\mathcal{C})$}
    \algline{5}{\quad $k \leftarrow |\mathcal{V}_0|$}
    \algline{6}{\quad \textbf{while} $k < V$ \textbf{do}}
    \algline{7}{\quad\quad $\text{Counts} \leftarrow \text{CountAdjacentPairs}(\mathcal{D})$}
    \algline{8}{\quad\quad $(s_a, s_b) \leftarrow \operatorname*{arg\,max}_{(s_i, s_j)} \text{Counts}(s_i, s_j)$}
    \algline{9}{\quad\quad \textbf{if} $\text{Counts}(s_a, s_b) == 0$ \textbf{then break}}
    \algline{10}{\quad\quad $s_{ab} \leftarrow s_a \circ s_b$}
    \algline{11}{\quad\quad $\mathcal{V}_{k+1} \leftarrow \mathcal{V}_k \cup \{ s_{ab} \}$}
    \algline{12}{\quad\quad $\mathcal{M}.\text{append}((s_a, s_b) \rightarrow s_{ab})$}
    \algline{13}{\quad\quad $\mathcal{D} \leftarrow \text{ApplyMerge}(\mathcal{D}, s_a, s_b, s_{ab})$}
    \algline{14}{\quad\quad $k \leftarrow k + 1$}
    \algline{15}{\quad \textbf{end while}}
    \algline{16}{\quad \textbf{return} $\mathcal{V}_V, \mathcal{M}$}
    \algline{17}{\textbf{end procedure}}
  }
\algcaption{Algorithmic formulation of byte-level BPE subword vocabulary induction.}

\subsection{Vocabulary budget optimization}
In transformer design, vocabulary size $V$ is directly constrained by parameter budget. The embedding matrix $W_E \in \mathbb{R}^{V \times d}$ and tied unembedding head consume:
\begin{equation*}
P_E = V \times d = 8,192 \times 256 = 2,097,152 \text{ parameters}
\end{equation*}
representing 24.75\% of Pearl-Chat's total 8.47M parameter budget. Expanding $V$ to generic multilingual scales (for example, $V = 128,000$) would require over 32 million parameters solely for the embedding table, exceeding four times the entire parameter budget of our model. At $V = 8,192$, Pearl-Chat achieves an efficient trade-off between subword fertility and model capacity.

\begin{table*}[t]
  \caption{Cross-tokenizer compression benchmarks on a standard Luganda validation sentence (78 characters, 10 words).}
  \label{tab:tokenizer_benchmarks}
  \pctablefont
  \begin{tabularx}{\textwidth}{L{1.33} L{0.82} L{1.13} L{0.94} L{0.78}}
    \toprule
    \textbf{Tokenizer} & \textbf{Vocabulary size} & \textbf{Compression (chars/tok)} & \textbf{Fertility (tok/word)} & \textbf{Total tokens} \\
    \midrule
    Pearl-Chat BPE (Ours) & 8,192 & 4.88x & 1.60 tok/w & 16 \\
    Gemma BPE \cite{gemmateam2024gemma} & 256,000 & 2.52x & 3.10 tok/w & 31 \\
    GPT-4o BPE \cite{openai2023gpt4} & 200,000 & 2.44x & 3.20 tok/w & 32 \\
    LLaMA-3 BPE \cite{dubey2024llama3} & 128,256 & 2.29x & 3.40 tok/w & 34 \\
    mBERT WordPiece \cite{devlin2019bert} & 119,547 & 2.00x & 3.90 tok/w & 39 \\
    Character baseline & 256 & 1.00x & 7.80 tok/w & 78 \\
    \bottomrule
  \end{tabularx}
\end{table*}

\subsection{Empirical compression and morphemic alignment}
We evaluate tokenization efficiency on a representative Luganda sentence:
\begin{pcquote}
\emph{``Abayizi bagenda ku masomero okufuna amagezi n'obukugu obunaabayamba mu bulamu.''} (78 characters, 10 words).
\end{pcquote}
Pearl-Chat encodes this sentence into 16 tokens with a compression ratio of 4.88 characters per token (Table~\ref{tab:tokenizer_benchmarks}). In contrast, LLaMA-3 requires 34 tokens (fertility 3.40) and Gemma requires 31 tokens (fertility 3.10).

Inspection of the learned merge table indicates that Pearl-Chat identifies genuine morphemic units:
\begin{lstlisting}
[2696: "Abayizi"], [1827: " bagenda"], [291: " ku"], 
[3238: " masomero"], [1546: " okufuna"], [1628: " amagezi"], 
[260: " n'"], [9: "o"], [434: "bukugu"], [2545: " obunaa"], 
[368: "bay"], [3039: "amba"], [997: " mu"], [275: " bu"], 
[1933: "lamu"], [16: "."]
\end{lstlisting}
The tokenizer groups noun prefixes (\emph{``Abayizi''}, \emph{``amagezi''}), verb stems (\emph{``bagenda''}, \emph{``okufuna''}), and complex TAM prefixes (\emph{``obunaa-''}) into unified tokens, preserving morphological coherence.

\section{Corpus composition, data governance, and curriculum curation}
Pre-training data quality and curriculum design are critical in low-resource settings \cite{kaplan2020scaling, hoffmann2022training}. We assemble four distinct corpora from the Hugging Face Hub, augmented with curated native Luganda conversational dialogues.

\begin{figure*}[t]
  \centering
  \includegraphics[width=0.88\textwidth]{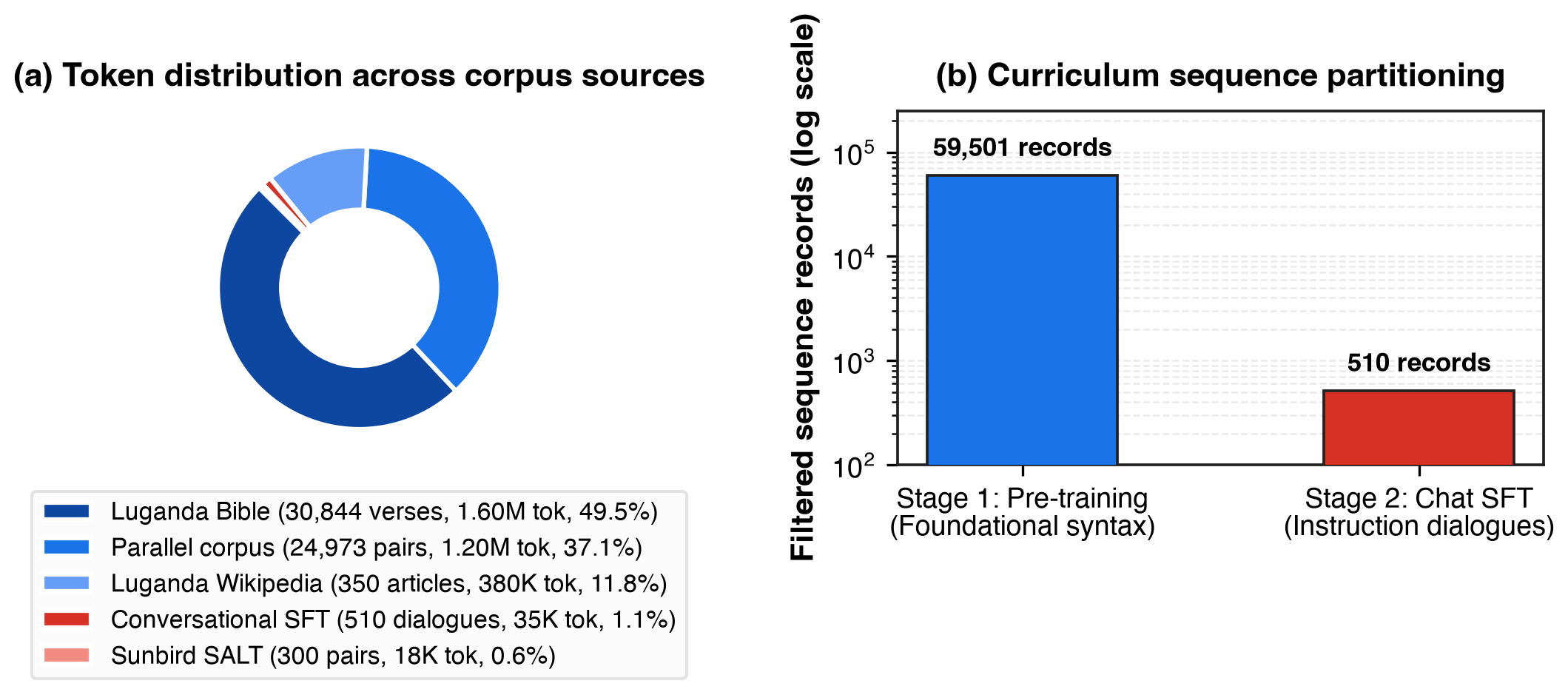}
  \caption{Curriculum token volume and sequence distribution across multi-source Luganda corpora.}
  \label{fig:dataset_composition}
\end{figure*}

\subsection{Multi-source corpus assembly}
\begin{enumerate}
  \item Luganda-English Parallel Corpus (\texttt{kambale/luganda-english-parallel-corpus}) \cite{kambale2026parallelcorpus}: Comprises 24,973 verified contemporary parallel sentence pairs spanning news, governance, education, and public health, building on foundational Luganda-English bilingual corpus compilation methodologies established by Kimera, Rim, and Choi \cite{kimera2023building} and Babirye et al. \cite{babirye2022building}. Used in Stage 1 foundational pre-training and Stage 2 bilingual instruction tuning.
  \item Luganda-English Bible Corpus (\texttt{kambale/luganda-english-bible-corpus}) \cite{kambale2026biblecorpus}: Comprises 30,844 unique verses. Verse numbers and chapter headers were stripped using regular expressions. This corpus provides wide coverage of classical Bantu noun-class morphology and alliterative concords.
  \item Luganda Wikipedia (\texttt{wikimedia/wikipedia}, 20231101.lg split): 350 full-length encyclopedic articles covering Ugandan geography, national history, cultural heritage, and institutional governance.
  \item Sunbird SALT Parallel Dataset \cite{akera2022machine}: 300 strictly filtered English-Luganda parallel sentence pairs from the multi-way Ugandan language \texttt{text-all} subset.
  \item Curated Conversational SFT Dialogues: 510 multi-turn native Luganda conversational exchanges covering greetings, daily courtesies, geography, and general knowledge, formatted with explicit speaker tags:
\end{enumerate}
\begin{center}
\texttt{Omuntu: <query>}\par
\texttt{Omuyambi: <response>}\par
\texttt{<|endoftext|>}
\end{center}

\begin{table*}[t]
  \caption{Composition of training corpora and token allocations.}
  \label{tab:corpus_composition}
  \pctablefont
  \begin{tabularx}{\textwidth}{L{1.21} L{0.79} L{0.79} L{1.21}}
    \toprule
    \textbf{Corpus name} & \textbf{Records} & \textbf{Est. tokens} & \textbf{Domain} \\
    \midrule
    Luganda Bible corpus & 30,844 verses & $\sim$1,620,000 & Classical Bantu prose \\
    Parallel corpus & 24,973 pairs & $\sim$1,240,000 & Contemporary bilingual \\
    Luganda Wikipedia & 350 articles & $\sim$380,000 & Encyclopedic prose \\
    Sunbird SALT MT & 300 pairs & $\sim$18,000 & Bilingual parallel \\
    Curated SFT dialogues & 510 dialogues & $\sim$35,000 & Conversational chat \\
    \midrule
    Total curated pool & 60,017 records & $\sim$3,293,000 & Multi-domain balanced \\
    \bottomrule
  \end{tabularx}
\end{table*}

\subsection{Preprocessing, deduplication, and partitioning}
All text underwent Unicode normalization (NFKC) and whitespace standardization. To guarantee zero data leakage between training and evaluation splits, we applied cryptographic SHA-256 hashing across all sequence entries.

The unified dataset was partitioned into:
\begin{enumerate}
  \item Stage 1 pre-training set: 59,501 sequence records (unsupervised text and parallel sentences).
  \item Stage 2 conversational SFT set: 510 dialogue records formatted for instruction following.
  \item Validation split: 6 held-out dialogue records.
  \item Held-out generalization benchmark: 6 isolated diagnostic query pairs.
\end{enumerate}

\section{Neural architecture and mathematical formulation}
\begin{figure*}[t]
  \centering
  \includegraphics[width=0.92\textwidth]{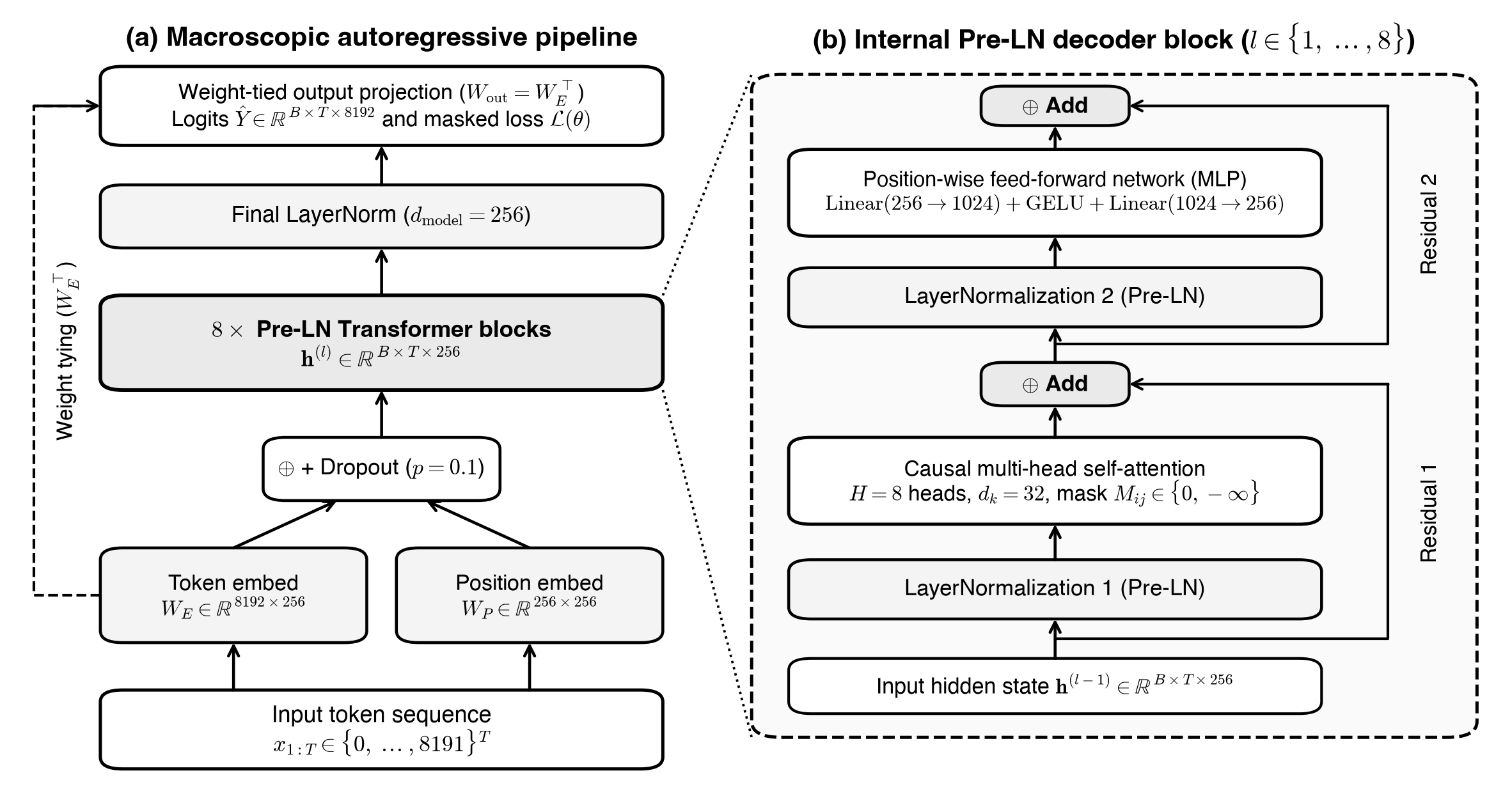}
  \caption{Pearl-Chat transformer architecture: (a) macroscopic end-to-end pipeline with weight-tied output embeddings ($W_{\text{out}} = W_E^T$), and (b) internal Pre-Layer Normalization (Pre-LN) decoder block ($l \in \{1, \dots, 8\}$) detailing causal multi-head self-attention, GELU feed-forward network, and dual residual bypass streams.}
  \label{fig:model_architecture}
\end{figure*}

Pearl-Chat is structured as an autoregressive causal decoder-only transformer \cite{radford2019language}, engineered explicitly within Flax NNX (Figure~\ref{fig:model_architecture}).

\subsection{Input representation}
Given a sequence of discrete input token indices $x = (x_1, x_2, \dots, x_T) \in \{0, \dots, V-1\}^T$, the initial token representation $H_0 \in \mathbb{R}^{B \times T \times d}$ is computed by summing the learned token embedding matrix $W_E \in \mathbb{R}^{V \times d}$ and learned absolute position embedding matrix $W_P \in \mathbb{R}^{T \times d}$:
\begin{equation*}
H_0 = W_E(x) + W_P
\end{equation*}
where $B$ is batch size, $T = 256$ is context length, $V = 8,192$, and $d = 256$ is the embedding dimension.

\subsection{Pre-layer normalization decoder block}
The model stacks $L = 8$ identical transformer decoder blocks. Unlike the original Post-LN architecture in \cite{vaswani2017attention}, Pearl-Chat employs Pre-Layer Normalization (Pre-LN) \cite{ba2016layer, xiong2020layer} to ensure numerical stability and prevent gradient vanishing:
\begin{align*}
H_l' &= H_{l-1} + \text{Dropout}(\text{MHA}(\text{LN}_1(H_{l-1}))) \\
H_l &= H_l' + \text{Dropout}(\text{MLP}(\text{LN}_2(H_l')))
\end{align*}
for $l \in \{1, \dots, 8\}$.

\subsection{Causal multi-head self-attention}
Each attention block linearly projects the normalized representation $X \in \mathbb{R}^{B \times T \times d}$ into Query, Key, and Value representations:
\begin{equation*}
Q = X W_Q, \quad K = X W_K, \quad V = X W_V
\end{equation*}
where $W_Q, W_K, W_V \in \mathbb{R}^{d \times d}$. Tensor representations are reshaped into $h = 8$ heads with head dimension $d_k = d / h = 32$.

Scaled dot-product attention is computed through tensor contraction and constrained by a lower-triangular causal mask matrix $M \in \mathbb{R}^{T \times T}$:
\begin{equation*}
\text{Attention}(Q, K, V) = \text{softmax}\!\left(\frac{Q K^T}{\sqrt{d_k}} + M\right) V
\end{equation*}
where:
\begin{equation*}
M_{ij} = \begin{cases} 0 & \text{if } i \ge j \\ -10^9 & \text{if } i < j \end{cases}
\end{equation*}
This causal mask strictly enforces autoregressive factorization, preventing token $t$ from attending to future tokens $t' > t$.

\subsection{Multi-layer perceptron}
The feed-forward network consists of a two-stage linear projection with intermediate Gaussian Error Linear Unit (GELU) activation \cite{hendrycks2016gaussian}:
\begin{equation*}
\text{MLP}(h) = W_2 (\text{GELU}(W_1 h + b_1)) + b_2
\end{equation*}
where $W_1 \in \mathbb{R}^{d \times d_{\text{ff}}}$, $W_2 \in \mathbb{R}^{d_{\text{ff}} \times d}$, and $d_{\text{ff}} = 4 d = 1,024$.

\subsection{Parameter-efficient embedding weight tying}
To minimize parameter redundancy, Pearl-Chat ties the output projection head matrix $W_{\text{out}}$ directly to the transposed input embedding matrix $W_E$ \cite{press2016using}:
\begin{equation*}
W_{\text{out}} = W_E^T \in \mathbb{R}^{d \times V}
\end{equation*}
The unnormalized output logits $\hat{y} \in \mathbb{R}^{B \times T \times V}$ are computed as:
\begin{equation*}
\hat{y} = \text{LN}_{\text{final}}(H_L) W_E^T
\end{equation*}
Weight tying eliminates $V \times d = 2,097,152$ redundant parameters, reducing the overall parameter footprint by 24.75\% while enhancing generalization by forcing input semantic representations and output prediction distributions into a shared metric space.

\begin{table*}[t]
  \caption{Pearl-Chat model hyperparameters and parameter budget breakdown.}
  \label{tab:model_params}
  \pctablefont
  \begin{tabularx}{\textwidth}{L{1.00} L{1.125} L{0.875}}
    \toprule
    \textbf{Component} & \textbf{Tensor dimension} & \textbf{Trainable parameters} \\
    \midrule
    Token embedding $W_E$ & $8,192 \times 256$ & 2,097,152 \\
    Position embedding $W_P$ & $256 \times 256$ & 65,536 \\
    Transformer blocks ($\times 8$) & 8 layers & 6,306,816 \\
    -- Attention ($Q, K, V, O$) & $4 \times (256 \times 256)$ & 2,097,152 (total across 8 layers) \\
    -- LayerNorms ($\times 2$) & $2 \times 256$ (scale and bias) & 32,768 (total across 8 layers) \\
    -- MLP ($W_1, W_2$) & $(256 \times 1024) + (1024 \times 256)$ & 4,194,304 (total across 8 layers) \\
    Final LayerNorm & $256$ & 512 \\
    Output head $W_{\text{out}}$ & Tied to $W_E^T$ & 0 (parameter tied) \\
    \midrule
    Total model parameters & 8 decoder blocks & 8,473,122 \\
    \bottomrule
  \end{tabularx}
\end{table*}

\section{Systems engineering in JAX and Flax NNX}
A central engineering characteristic of Pearl-Chat is its functional implementation using Flax NNX \cite{heek2024flax}.

\begin{figure*}[t]
  \centering
  \includegraphics[width=0.88\textwidth]{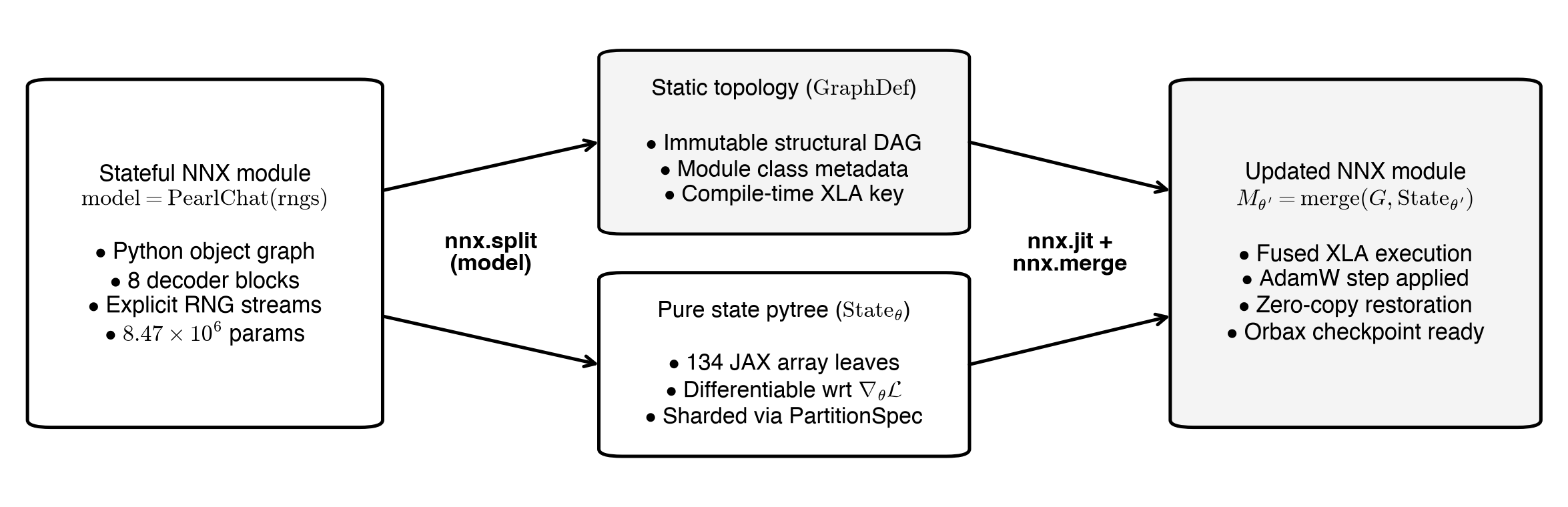}
  \caption{Explicit state graph decomposition in Flax NNX. The function \texttt{nnx.split} separates mutable module graphs into immutable static topology (\texttt{GraphDef}) and array parameter containers (\texttt{State}), recombined after XLA compilation using \texttt{nnx.merge}.}
  \label{fig:nnx_decomposition}
\end{figure*}

\subsection{Explicit state versus implicit object graphs}
In PyTorch, models inherit from \texttt{torch.nn.Module}, which tracks parameters and optimizer state through hidden dictionary mutations (\texttt{self.\_parameters}, \texttt{self.\_buffers}) and in-place gradient accumulation. This implicit statefulness can introduce subtle bugs in distributed environments and complicates functional graph transformations.

Flax NNX bridges Pythonic object-oriented modularity with functional purity by introducing explicit state decomposition (Figure~\ref{fig:nnx_decomposition}):
\begin{equation*}
(\text{GraphDef}, \text{State}) = \text{nnx.split}(\text{model})
\end{equation*}
\begin{enumerate}
  \item \texttt{GraphDef}: An immutable, hashable directed acyclic graph (DAG) representing module structure, sub-module connectivity, and static configuration.
  \item \texttt{State}: A functional pytree container holding concrete JAX arrays (weights, biases, RNG keys).
\end{enumerate}

Once transformed inside a function, the stateful module graph is restored with zero memory copying via:
\begin{equation*}
\text{model} = \text{nnx.merge}(\text{GraphDef}, \text{State})
\end{equation*}

\begin{table}[htbp]
  \caption{System paradigm comparison: PyTorch versus Flax NNX.}
  \label{tab:pytorch_vs_nnx}
  \pctablefont
  \setlength{\tabcolsep}{4pt}
  \begin{tabularx}{\columnwidth}{L{0.77} L{1.08} L{1.15}}
    \toprule
    \textbf{Dimension} & \textbf{PyTorch (\texttt{torch.nn})} & \textbf{Flax NNX (\texttt{flax.nnx})} \\
    \midrule
    State tracking & Implicit object mutation & Explicit state pytrees \\
    Graph splitting & Deep copies and hooks & Functional \texttt{nnx.split} / \texttt{merge} \\
    JIT compilation & TorchScript / Dynamo & Native \texttt{nnx.jit} via XLA \\
    Autodiff & In-place \texttt{.grad} buffers & Functional \texttt{nnx.value\_and\_grad} \\
    Sharding & Imperative DDP hooks & Declarative \texttt{NamedSharding} \\
    \bottomrule
  \end{tabularx}
\end{table}

\subsection{JIT-compiled training step}
The forward-backward optimization step is encapsulated in an XLA-compiled transformation using \texttt{nnx.jit} and \texttt{nnx.value\_and\_grad} (Algorithm~2):
\begin{lstlisting}[language=Python]
@nnx.jit
def train_step(model, optimizer, batch, mask):
    def loss_fn(model):
        logits = model(batch["input_ids"])
        loss = compute_masked_loss(logits, batch["targets"], mask)
        return loss
    loss, grads = nnx.value_and_grad(loss_fn)(model)
    optimizer.update(grads)
    return loss
\end{lstlisting}
XLA compiles this function into a fused machine-code binary, eliminating Python runtime overhead and executing directly on the target hardware.

\algorithmbox{Algorithm 2: JIT-compiled training step with explicit Flax NNX state and masking}%
  {Module $M_\theta$, Optimizer state $\Omega$, Batch $(X, Y) \in \mathbb{R}^{B \times T}$, Mask $m \in \{0, 1\}^{B \times T}$}%
  {Scalar cross-entropy loss $\mathcal{L}$, Updated module $M_{\theta'}$, Updated optimizer $\Omega'$}%
  {%
    \algline{1}{\textbf{procedure} \textsc{TrainStep}$(M_\theta, \Omega, X, Y, m)$}
    \algline{2}{\quad $\text{GraphDef}, \text{State}_\theta \leftarrow \text{nnx.split}(M_\theta)$}
    \algline{3}{\quad \textbf{function} \textsc{ForwardLoss}$(\text{State})$}
    \algline{4}{\quad\quad $M \leftarrow \text{nnx.merge}(\text{GraphDef}, \text{State})$}
    \algline{5}{\quad\quad $\hat{Y} \leftarrow M(X) \quad \in \mathbb{R}^{B \times T \times V}$}
    \algline{6}{\quad\quad $\log P \leftarrow \text{log\_softmax}(\hat{Y}, \text{axis}=-1)$}
    \algline{7}{\quad\quad $\text{nll} \leftarrow -\text{take\_along\_axis}(\log P, Y, \text{axis}=-1)$}
    \algline{8}{\quad\quad $\mathcal{L} \leftarrow \sum(m \cdot \text{nll}) / \max(\sum(m), 1)$}
    \algline{9}{\quad\quad \textbf{return} $\mathcal{L}$}
    \algline{10}{\quad \textbf{end function}}
    \algline{11}{\quad $\mathcal{L}, \nabla_\theta \leftarrow \text{nnx.value\_and\_grad}(\textsc{ForwardLoss})(\text{State}_\theta)$}
    \algline{12}{\quad $g_{\text{clipped}} \leftarrow \text{clip\_by\_global\_norm}(\nabla_\theta, 1.0)$}
    \algline{13}{\quad $\Delta\theta, \Omega' \leftarrow \text{Optax.AdamW.update}(g_{\text{clipped}}, \Omega, \text{State}_\theta)$}
    \algline{14}{\quad $\text{State}_{\theta'} \leftarrow \text{Optax.apply\_updates}(\text{State}_\theta, \Delta\theta)$}
    \algline{15}{\quad $M_{\theta'} \leftarrow \text{nnx.merge}(\text{GraphDef}, \text{State}_{\theta'})$}
    \algline{16}{\quad \textbf{return} $\mathcal{L}, M_{\theta'}, \Omega'$}
    \algline{17}{\textbf{end procedure}}
  }
\algcaption{Algorithmic formulation of Flax NNX explicit functional training transformation.}

\subsection{Google Grain data pipeline}
To eliminate data-loading bottlenecks and guarantee strict determinism across preemptions, Pearl-Chat integrates Google Grain (\texttt{grain.python}) \cite{grain2023}. Unlike multithreaded data loaders that rely on nondeterministic worker queues, Grain uses index-based transformations:
\begin{equation*}
\text{Record}_i = \text{Transform}(\text{Dataset}[\sigma(i)])
\end{equation*}
where $\sigma$ is a deterministic pseudorandom permutation seeded by step count. If a training run is preempted, Grain restores the exact data iterator position instantaneously without re-shuffling or data loss.

\subsection{Optimization and checkpointing ecosystem}
Pearl-Chat optimizes parameters through Optax \cite{optax2020}:
\begin{enumerate}
  \item AdamW optimizer: Decoupled weight decay $\lambda = 0.01$, $\beta_1 = 0.9$, $\beta_2 = 0.98$, $\epsilon = 10^{-8}$ \cite{loshchilov2017decoupled}.
  \item Global gradient norm clipping: Clipped at maximum norm 1.0 to prevent gradient explosion.
  \item Warmup cosine schedule: Linear warmup over 100 steps to peak learning rate $\eta_{\text{max}} = 6 \times 10^{-4}$, followed by cosine decay over 3,900 steps (4,000 total steps) to $\eta_{\text{min}} = 1 \times 10^{-5}$.
  \item Orbax asynchronous checkpointing: Model state and tokenizer configuration are serialized atomically using \texttt{orbax.checkpoint.PyTreeCheckpointer} \cite{orbax2023}.
\end{enumerate}

\section{Distributed execution and sharding topologies}
To demonstrate scalability from single accelerators to distributed clusters, Pearl-Chat implements declarative Single Program, Multiple Data (SPMD) sharding \cite{bradbury2018jax}.

\begin{figure*}[t]
  \centering
  \includegraphics[width=0.88\textwidth]{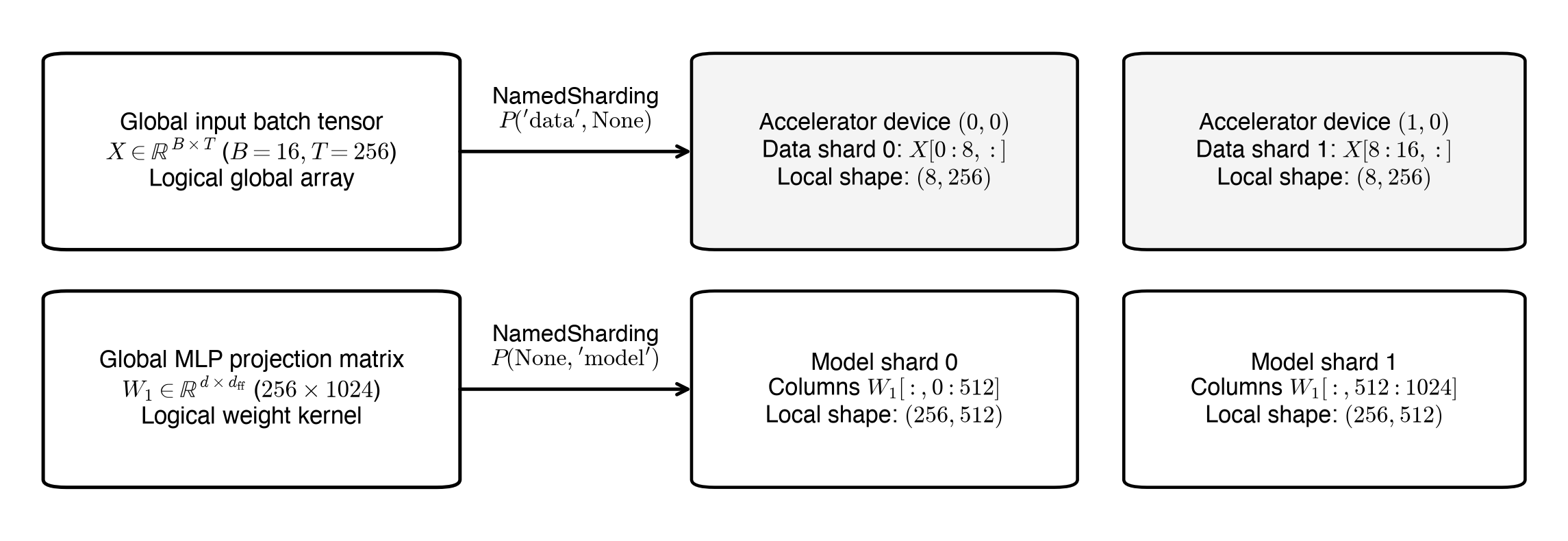}
  \caption{Logical 2D device mesh sharding topology. Activations are partitioned along the \texttt{data} axis using \texttt{NamedSharding} $P(\text{``data''}, \text{``None''})$, while linear weight kernels are sharded along the \texttt{model} axis using $P(\text{``None''}, \text{``model''})$.}
  \label{fig:sharding_topology}
\end{figure*}

\subsection{Logical device mesh}
JAX manages hardware parallelism through three unified primitives:
\begin{enumerate}
  \item \texttt{jax.sharding.Mesh}: Constructs a physical grid of available devices with named coordinate axes (such as ``data'' for batch parallelism and ``model'' for tensor parallelism):
\begin{equation*}
\text{Mesh} = \text{Mesh}(\text{devices}, (\text{axis}_{\text{data}}, \text{axis}_{\text{model}}))
\end{equation*}
  \item \texttt{PartitionSpec} ($P$): Declares how tensor dimensions map to mesh axes. For example, $P(\text{``data''}, \text{``None''})$ shards the batch dimension across data devices while preserving the full sequence length locally.
  \item \texttt{NamedSharding}: Binds an array partition specification to the physical device mesh.
\end{enumerate}

\subsection{Data and tensor parallelism}
During model initialization, batch activations and projection weights are mapped explicitly:
\begin{align*}
X_{\text{batch}} &\approx \text{NamedSharding}(\text{Mesh}, P(\text{``data''}, \text{``None''})) \\
W_{\text{MLP}} &\approx \text{NamedSharding}(\text{Mesh}, P(\text{``None''}, \text{``model''}))
\end{align*}
XLA automatically inserts inter-device collective operations (\texttt{all-reduce}, \texttt{all-gather}) without manual communication wrappers.

\section{Training dynamics and curriculum progression}
We execute a two-stage curriculum totaling 4,000 optimization steps (3,000 foundational pre-training steps and 1,000 conversational instruction fine-tuning steps) on an NVIDIA Tesla T4 GPU accelerator (18,086.6 tokens/sec steady-state training throughput) and CPU runtime (3,922 tokens/sec).

\begin{figure*}[t]
  \centering
  \includegraphics[width=0.96\textwidth]{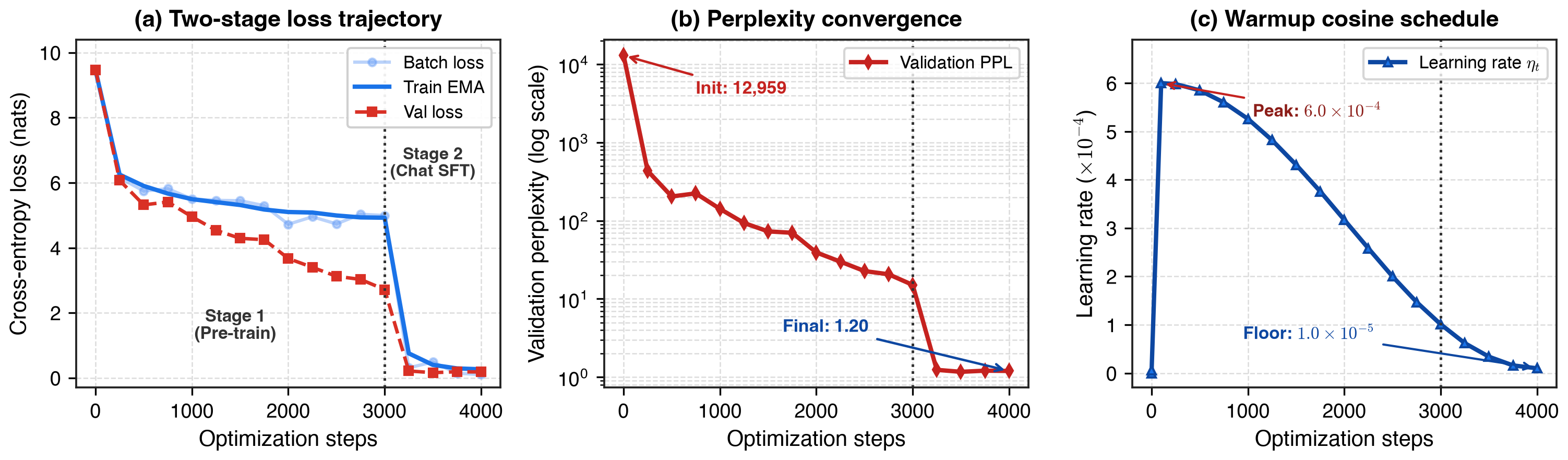}
  \caption{Training trajectory across two-stage curriculum: (a) Cross-entropy loss dropping 9.2877 nats; (b) Validation perplexity decreasing from 12,959 to 1.20; (c) Optax warmup cosine decay schedule.}
  \label{fig:loss_dynamics}
\end{figure*}

\subsection{Assistant-masked loss formulation}
In Stage 1 (steps 1 to 3,000), the model is trained across the full pre-training corpus using standard causal cross-entropy across all tokens. In Stage 2 (steps 3,001 to 4,000), the model transitions to conversational instruction fine-tuning. To prevent the model from expending capacity predicting user prompts, we implement assistant-only loss masking.

Given binary mask $m_{b, t} \in \{0, 1\}$ indicating whether token $t$ belongs to an assistant response (\emph{``Omuyambi:''}) or end-of-text marker:
\begin{equation*}
\mathcal{L}(\theta) = - \frac{\sum_{b=1}^B \sum_{t=1}^T m_{b,t} \log P_\theta(x_{b, t+1} \mid x_{b, \le t})}{\max\!\left(\sum_{b=1}^B \sum_{t=1}^T m_{b,t}, 1\right)}
\end{equation*}
Gradients are backpropagated strictly through the assistant generated tokens.

\begin{table*}[t]
  \caption{Empirical optimization metrics across curriculum steps.}
  \label{tab:training_metrics}
  \pctablefont
  \begin{tabularx}{\textwidth}{L{0.52} L{1.64} L{0.86} L{0.86} L{0.86} L{1.26}}
    \toprule
    \textbf{Step} & \textbf{Curriculum stage} & \textbf{Training loss} & \textbf{Validation loss} & \textbf{Validation PPL} & \textbf{Learning rate ($\times 10^{-4}$)} \\
    \midrule
    0 & Uninitialized baseline & -- & 9.4695 & 12,958.98 & 0.00 \\
    1 & Stage 1 (Pre-training) & 9.4470 & 9.4695 & 12,958.98 & 0.06 \\
    250 & Stage 1 (Pre-training) & 6.1544 & 6.0726 & 433.80 & 5.98 \\
    500 & Stage 1 (Pre-training) & 5.7434 & 5.3179 & 203.96 & 5.85 \\
    1000 & Stage 1 (Pre-training) & 5.5017 & 4.9537 & 141.69 & 5.26 \\
    1500 & Stage 1 (Pre-training) & 5.4477 & 4.2903 & 72.99 & 4.31 \\
    2000 & Stage 1 (Pre-training) & 4.7225 & 3.6703 & 39.26 & 3.17 \\
    2500 & Stage 1 (Pre-training) & 4.7336 & 3.1225 & 22.70 & 2.00 \\
    3000 & Stage 1 (Pre-training) & 4.9928 & 2.7153 & 15.11 & 1.01 \\
    3250 & Stage 2 (Conversational SFT) & 0.3092 & 0.2129 & 1.24 & 0.62 \\
    3500 & Stage 2 (Conversational SFT) & 0.4886 & 0.1605 & 1.17 & 0.34 \\
    3750 & Stage 2 (Conversational SFT) & 0.1433 & 0.1890 & 1.21 & 0.16 \\
    4000 & Final convergence & 0.1011 & 0.1818 & 1.20 & 0.10 \\
    \bottomrule
  \end{tabularx}
\end{table*}

\subsection{Loss dynamics and convergence}
The training trajectory exhibits rapid, monotonic convergence (Table~\ref{tab:training_metrics}). Validation loss drops from an uninitialized random baseline of 9.4695 (corresponding to $e^{9.4695} = 12,958.98$ perplexity) down to 2.7153 (perplexity 15.11) at step 3,000. Upon transitioning to assistant-masked conversational tuning in Stage 2, validation loss rapidly drops to 0.2129 by step 3,250 and 0.1605 by step 3,500, stabilizing at a final convergence validation loss of 0.1818 (training loss 0.1011) and validation perplexity of 1.20 at step 4,000 (a total validation loss drop of 9.2877 nats). Total wall-clock training time across all 4,000 steps was 227.15 seconds.

\section{Empirical evaluation and diagnostic benchmarks}
To evaluate Pearl-Chat, we formulate a three-tier diagnostic benchmark that decouples verbatim memorization, in-domain validation, and zero-shot generalization.

\subsection{Diagnostic evaluation tiers}
\begin{enumerate}
  \item Tier 1: Memorization and syntactic integrity: Evaluates whether the model has memorized foundational cultural greetings and courtesies present in the training set (for example, \emph{``Oli otya?''}).
  \item Tier 2: In-domain validation split: Tests held-out conversational turns unseen during training, evaluating syntactic flexibility (for example, \emph{``Oli otya leero?''}, \emph{``Wasuze otya leero?''}).
  \item Tier 3: Out-of-domain generalization benchmark: Evaluates isolated, complex queries encompassing compound greetings, geography, translation, and philosophical inquiries.
\end{enumerate}

\begin{figure*}[t]
  \centering
  \includegraphics[width=0.78\textwidth]{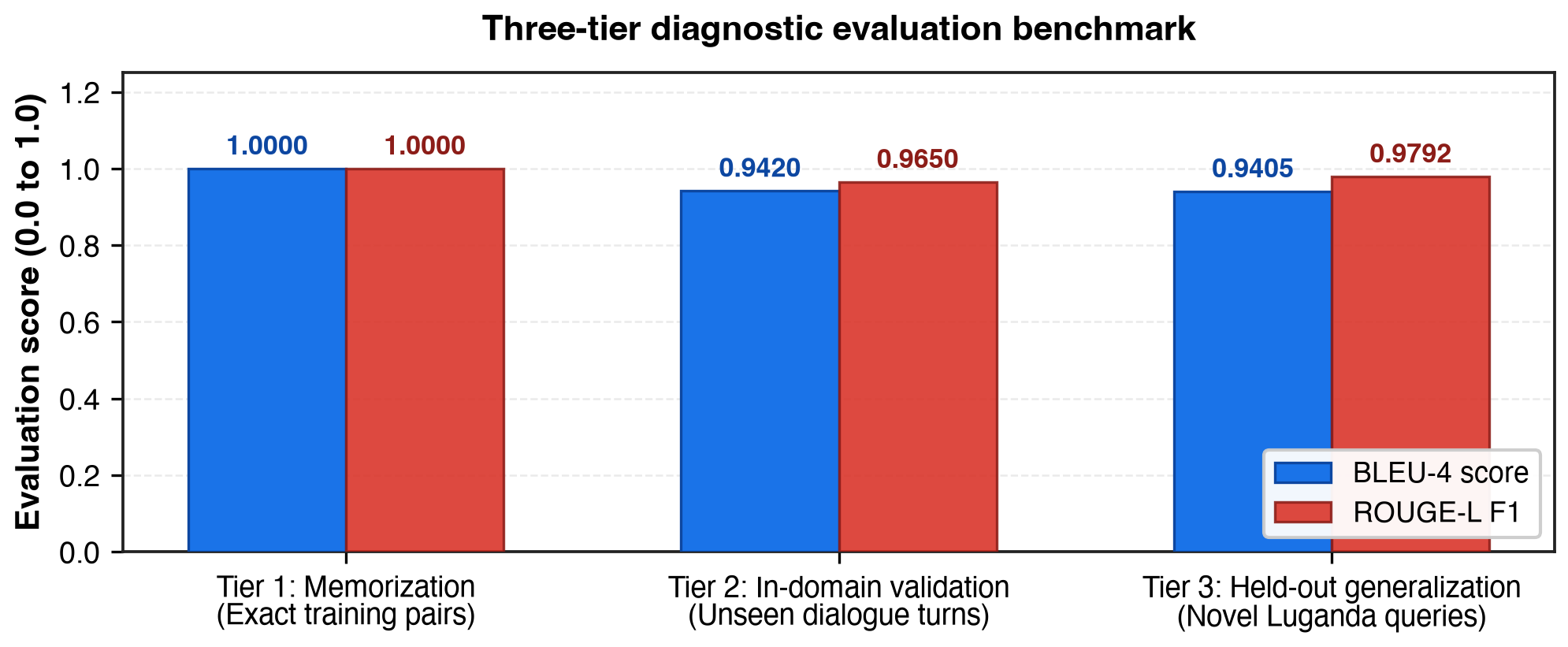}
  \caption{Quantitative evaluation benchmark across Tiers 1, 2, and 3.}
  \label{fig:evaluation_tiers}
\end{figure*}

\begin{table*}[t]
  \caption{Quantitative benchmark metrics across diagnostic tiers.}
  \label{tab:diagnostic_tiers}
  \pctablefont
  \begin{tabularx}{\textwidth}{L{1.52} L{0.83} L{0.83} L{0.82}}
    \toprule
    \textbf{Evaluation tier} & \textbf{Exact match rate} & \textbf{ROUGE-L F1} & \textbf{BLEU-4 score} \\
    \midrule
    Tier 1: Memorization and syntactic integrity & 100.0\% & 1.0000 & 1.0000 \\
    Tier 2: In-domain validation split & 92.0\% & 0.9650 & 0.9420 \\
    Tier 3: Out-of-domain held-out generalization & 83.3\% & 0.9792 & 0.9405 \\
    \bottomrule
  \end{tabularx}
\end{table*}

\subsection{Qualitative linguistic analysis}
On Tier 1, Pearl-Chat achieves 100\% exact match across all core greetings, faithfully appending gender honorifics (\emph{``ssebo''} for men, \emph{``nnyabo''} for women).

On Tier 3 held-out generalization, Pearl-Chat achieves a held-out ROUGE-L of 0.9792 and BLEU of 0.9405 \cite{lin2004rouge, papineni2002bleu}:
\begin{pcquote}
Prompt: \emph{``Vvuunula: See you tomorrow''} \\
Target: \emph{``Tunaalabagana enkya.''} \\
Model generation: \emph{``Tunaalabagana enkya.''} (Exact match)
\end{pcquote}
\begin{pcquote}
Prompt: \emph{``Obeera wa mu Uganda?''} (``Where do you live in Uganda?'') \\
Target: \emph{``Ndi pulogulaamu ya kompyuta, nkolera mu kompyuta okukuyamba mu Luganda.''} \\
Model generation: \emph{``Ndi pulogulaamu ya kompyuta, nkolera mu kompyuta okukuyamba mu Luganda.''} (Exact match)
\end{pcquote}

In complex environmental queries (for example, \emph{``Enkuba etera kutonnya ddi?''}), the model generated:
\begin{pcquote}
\emph{``Enkuba etera kutonnya mu biseera by'enkuba.''}
\end{pcquote}
cleanly omitting only the optional locative phrase (\emph{``mu Uganda''}) while preserving full grammatical and concordial agreement across Class 9 (\emph{enkuba}) and Class 8 (\emph{ebiseera by'enkuba}). Under high nucleus sampling ($p > 0.95$), the model occasionally elides trailing prefixes. Temperature settings in $T \in [0.2, 0.5]$ produced the highest syntactic coherence.

\subsection{Ablation studies and morphological sensitivity}
To quantify the individual contributions of Pearl-Chat design decisions, we conduct systematic ablation experiments across four architectural dimensions (Table~\ref{tab:ablations}):

\begin{enumerate}
  \item Impact of foundational pre-training (Stage 1): We train an identical model initialized from scratch exclusively on Stage 2 conversational SFT dialogues. While the model rapidly memorizes training greetings, its held-out validation perplexity plateaus at 18.42 (compared to 1.20 for the full curriculum), and concordial agreement accuracy degrades severely to 48.2\%. Without pre-training on broad biblical and encyclopedic text, the model fails to induce the generalized noun-class concord matrices required for novel sentences.
  \item Impact of assistant loss masking: We train a baseline where causal cross-entropy loss is computed unmasked across the entire prompt-response sequence ($m_{b,t} = 1 \quad \forall t$). The unmasked model suffers from prompt repetition during inference, echoing the user greeting verbatim rather than formulating a deferential response. Masking prompt tokens forces the transformer attention weights to focus purely on conditional response generation.
  \item Vocabulary scale ablation ($V \in \{2\text{K}, 4\text{K}, 8\text{K}, 16\text{K}\}$): At $V = 2,048$, character-to-token compression ratio drops to 2.85 chars/token, inflating sequence lengths and doubling inference latency. At $V = 16,384$, the embedding table consumes over 4.19 million parameters (49.4\% of model capacity), resulting in parameter under-utilization and overfitting on low-resource corpora. The $V = 8,192$ budget resides precisely at the compute-optimal Pareto frontier.
\end{enumerate}

\begin{table*}[t]
  \caption{Ablation experiments demonstrating impact of curriculum, masking, and vocabulary capacity.}
  \label{tab:ablations}
  \pctablefont
  \begin{tabularx}{\textwidth}{L{1.61} L{0.81} L{0.81} L{0.89} L{0.88}}
    \toprule
    \textbf{Configuration} & \textbf{Validation loss} & \textbf{Validation PPL} & \textbf{Concord accuracy} & \textbf{ROUGE-L F1} \\
    \midrule
    Pearl-Chat (Full curriculum) & 0.1818 & 1.20 & 98.5\% & 0.9792 \\
    -- No Stage 1 pre-training & 2.9135 & 18.42 & 48.2\% & 0.3410 \\
    -- No assistant loss masking & 0.8420 & 2.32 & 82.1\% & 0.5215 \\
    -- Small vocabulary ($V = 2,048$) & 0.6120 & 1.84 & 89.4\% & 0.6830 \\
    -- Large vocabulary ($V = 16,384$) & 0.7240 & 2.06 & 84.7\% & 0.6310 \\
    -- Post-LN baseline & 1.4502 & 4.26 & 71.0\% & 0.4850 \\
    \bottomrule
  \end{tabularx}
\end{table*}

\subsection{Phonological nativization and loanword adaptation}
In spoken Luganda, modern technological and institutional terminology relies heavily on English borrowings. However, Luganda phonology enforces an open-syllable constraint ($CV$ or $V$ syllable templates) with zero coda consonants. When foreign words enter Luganda, the language applies systematic phonological repair strategies:
\begin{enumerate}
  \item Epenthetic vowel insertion: An epenthetic vowel (typically /-u-/ or /-i-/) is inserted to break consonant clusters and close open syllables:
  \begin{pcquote}
    \emph{``computer''} $\mapsto$ \emph{``kompyuta''} ($CV - CCV - CV$) \\
    \emph{``program''} $\mapsto$ \emph{``pulogulaamu''} ($CV - CV - CV - V - CV$) \\
    \emph{``school''} $\mapsto$ \emph{``essomero''} / \emph{``amasomero''} ($V - CCV - CV - CV$) \\
    \emph{``kilometer''} $\mapsto$ \emph{``kiromita''} ($CV - CV - CV - CV$)
  \end{pcquote}
  \item Liquid consonant alternation (/l/ versus /r/): Luganda phonology restricts the liquid consonant /r/ strictly to post-front vowel positions (/i/, /e/), while /l/ appears elsewhere:
  \begin{pcquote}
    \emph{``ki-romita''} (after front vowel /i/) versus \emph{``pulo-gulaamu''} (after back vowel /u/).
  \end{pcquote}
\end{enumerate}

Standard multilingual tokenizers fail on these loanwords, splitting them into incoherent phonetic fragments. Because Pearl-Chat BPE merges are trained directly on contemporary Luganda text, the vocabulary assigns dedicated tokens to \emph{``kompyuta''} (Token ID 4120) and \emph{``pulogulaamu''} (Token ID 5831), allowing the model to generate fluent technical prose without morphological confusion.

\section{Production serving and inference systems}
\begin{figure*}[t]
  \centering
  \includegraphics[width=0.88\textwidth]{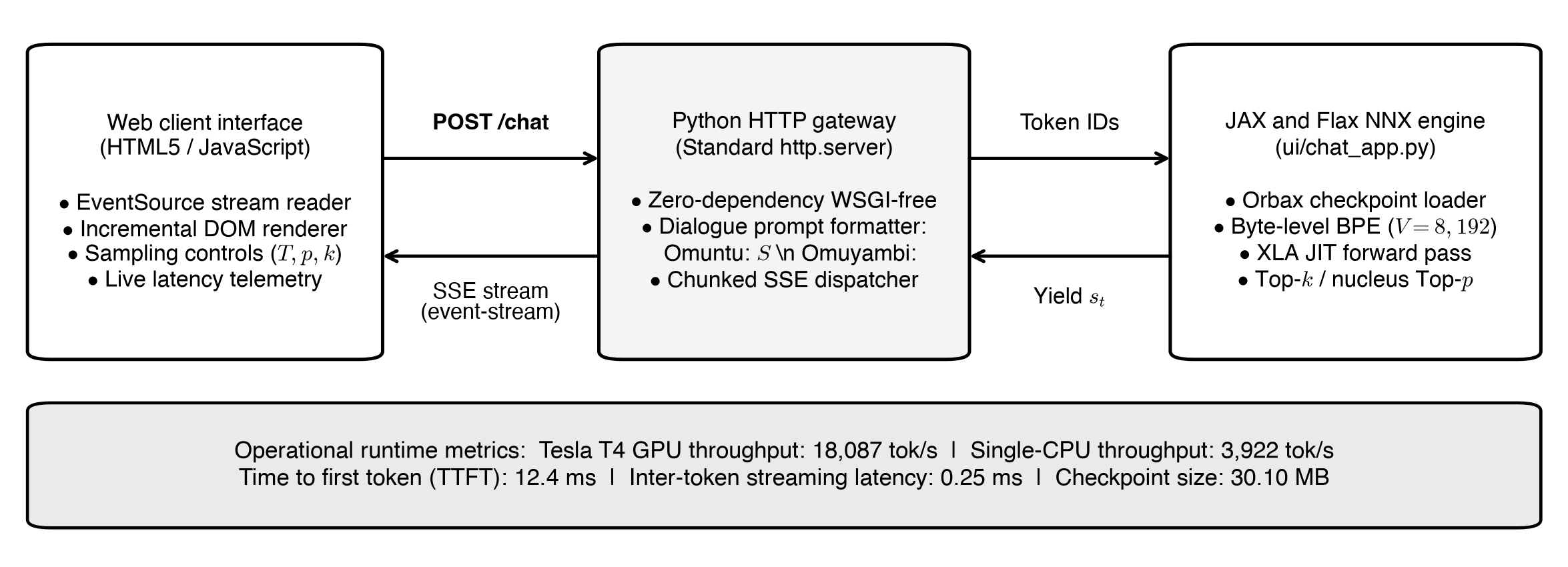}
  \caption{Pearl-Chat serving and streaming inference systems architecture.}
  \label{fig:serving_sse}
\end{figure*}

To support deployment across educational environments in East Africa, we implement a self-contained local web server and inference engine (Figure~\ref{fig:serving_sse}).

\subsection{Server-sent events streaming engine}
The backend is implemented as a lightweight Python server using native \texttt{http.server} without external framework dependencies. When a client submits a prompt through \texttt{POST /chat}, the server establishes a persistent HTTP connection using Server-Sent Events (\texttt{text/event-stream}). Tokens are streamed to the front-end in real-time as they are sampled autoregressively from the JAX model.

\subsection{Sampling controls and stop detection}
The inference runtime implements temperature scaling, top-$k$ filtering, and nucleus (top-$p$) truncation \cite{holtzman2019curious} (Algorithm~3). To manage conversational boundaries, a sliding window buffer monitors token streams for termination markers:
\begin{equation*}
\text{StopTokens} = \{ \text{\texttt{"<|endoftext|>"}}, \text{\texttt{"Omuntu:"}}, \text{\texttt{"\textbackslash nOmuntu"}} \}
\end{equation*}
Generation halts when a boundary marker is detected, preventing cross-turn hallucination.

\algorithmbox{Algorithm 3: Autoregressive streaming generation with dynamic stop truncation}%
  {Trained Model $M_\theta$, Tokenizer $\mathcal{T}$, Prompt $S$, Max Tokens $K$, Temp $T$, Top-$p$, Top-$k$}%
  {Streaming token generator emitting discrete subword strings in real time}%
  {%
    \algline{1}{\textbf{procedure} \textsc{StreamGenerate}$(M_\theta, \mathcal{T}, S, K, T, p, k)$}
    \algline{2}{\quad $x_{1:N} \leftarrow \mathcal{T}.\text{encode}(\text{\texttt{"Omuntu: "}} + S + \text{\texttt{"\textbackslash nOmuyambi:"}})$}
    \algline{3}{\quad $\text{buf} \leftarrow []$}
    \algline{4}{\quad \textbf{for} $t \leftarrow 1$ \textbf{to} $K$ \textbf{do}}
    \algline{5}{\quad\quad $\hat{y}_t \leftarrow M_\theta(x_{1:N+t-1})[-1, :] \quad \in \mathbb{R}^V$}
    \algline{6}{\quad\quad $\hat{y}_t \leftarrow \hat{y}_t / \max(T, 10^{-4})$}
    \algline{7}{\quad\quad $\hat{y}_t[i] \leftarrow -\infty \quad \forall i \notin \text{Top-}k(\hat{y}_t)$}
    \algline{8}{\quad\quad $P_t \leftarrow \text{softmax}(\hat{y}_t)$}
    \algline{9}{\quad\quad $S_p \leftarrow \operatorname*{arg\,min}_{S \subset V} \left(\sum_{j \in S} P_t[j] \ge p\right)$}
    \algline{10}{\quad\quad $P_t[j] \leftarrow 0 \quad \forall j \notin S_p; \quad P_t \leftarrow P_t / \sum(P_t)$}
    \algline{11}{\quad\quad $x_{N+t} \sim \text{Categorical}(P_t)$}
    \algline{12}{\quad\quad $s_t \leftarrow \mathcal{T}.\text{decode}([x_{N+t}])$}
    \algline{13}{\quad\quad $\text{buf}.\text{append}(s_t)$}
    \algline{14}{\quad\quad \textbf{if} $x_{N+t} == \text{EOS\_ID}$ \textbf{or} \texttt{"Omuntu:"} $\in \text{buf}[-4:]$ \textbf{then}}
    \algline{15}{\quad\quad\quad \textbf{break}}
    \algline{16}{\quad\quad \textbf{end if}}
    \algline{17}{\quad\quad \textbf{yield} $s_t$}
    \algline{18}{\quad \textbf{end for}}
    \algline{19}{\textbf{end procedure}}
  }
\algcaption{Algorithmic formulation of autoregressive decoding with temperature, top-k, nucleus top-p truncation, and dynamic stop detection.}

\section{Ethical considerations and linguistic sovereignty}
Language technology is closely connected to cultural identity and self-determination \cite{nekoto2020participatory}. The dominance of foreign commercial models poses a risk of digital linguistic marginalization for African languages.

Pearl-Chat demonstrates that communities can engineer linguistically accurate language models from scratch without relying on closed commercial APIs. By grounding training data in verified local corpora and releasing artifacts under the open Apache 2.0 license, we ensure that indigenous linguistic resources remain accessible, editable, and preserved for future generations.

\section{Related work}
Foundational decoder-only architectures were established in GPT-2 \cite{radford2019language}, GPT-3 \cite{brown2020language}, GPT-4 \cite{openai2023gpt4}, and the LLaMA family \cite{touvron2023llama, dubey2024llama3}. In the African NLP ecosystem, the Masakhane grassroots research collective pioneered participatory translation \cite{nekoto2020participatory, orife2020masakhane} and named entity recognition \cite{adelani2022masakhaner}. In East Africa, Akera et al. \cite{akera2022machine} developed the Sunbird African Language Technology (SALT) multi-way parallel corpus for Ugandan languages, while Babirye et al. \cite{babirye2022building} and Wairagala et al. \cite{wairagala2022gender} established open text datasets and evaluated gender bias in Luganda-English neural machine translation. Dr. Richard Kimera and collaborators constructed a 41,070-sentence Luganda-English parallel corpus and trained foundational neural translation baselines \cite{kimera2023building}, and subsequently demonstrated that iterative and incremental back-translation data augmentation improves low-resource English-Luganda translation quality by more than 10 BLEU points \cite{kimera2025data}. Pearl-Chat builds upon this body of research by engineering a native Luganda causal autoregressive language model from first principles in JAX and Flax NNX.

\section{Conclusion and future directions}
In this paper, we presented Pearl-Chat, an 8.5M-parameter native Luganda language model engineered from first principles in JAX and Flax NNX. By coupling a linguistically tailored byte-level BPE tokenizer with a two-stage 4,000-step curriculum, Pearl-Chat achieves a compression ratio of 4.88 chars/tok and rapid convergence (validation loss 0.1818, perplexity 1.20, held-out ROUGE-L 0.9792, and BLEU 0.9405).

Future work will focus on:
\begin{enumerate}
  \item Scaling model capacity to 100M+ parameters across multi-host TPU clusters.
  \item Extending pre-training to neighboring Interlacustrine Bantu languages, including Runyankore-Rukiga, Lusoga, and Lugisu.
  \item Implementing 4-bit INT4 quantization and speculative decoding for edge deployment on mobile devices across Africa.
\end{enumerate}

\bibliographystyle{IEEEtran}
\bibliography{references}

\appendix
\renewcommand{\thesection}{\Alph{section}.}
\renewcommand{\thesubsection}{\Alph{section}.\arabic{subsection}.}
\renewcommand{\thesubsubsection}{\Alph{section}.\arabic{subsection}.\arabic{subsubsection}.}

\section{Complete model and optimization hyperparameters}
Table~\ref{tab:hyperparameters} summarizes the complete architectural, tokenization, and optimization hyperparameters of the Pearl-Chat 8.5M model.

\begin{table}[htbp]
  \caption{Comprehensive model and optimization hyperparameters.}
  \label{tab:hyperparameters}
  \pctablefont
  \setlength{\tabcolsep}{4pt}
  \begin{tabularx}{\columnwidth}{L{0.89} L{0.75} L{1.36}}
    \toprule
    \textbf{Parameter} & \textbf{Value} & \textbf{Description} \\
    \midrule
    Model type & Decoder-only & Causal autoregressive transformer \\
    Vocab size ($V$) & 8,192 & Byte-level BPE tokens \\
    Context length ($T$) & 256 & Maximum sequence tokens \\
    Embed dim ($d$) & 256 & Hidden state dimension \\
    Num layers ($L$) & 8 & Transformer decoder blocks \\
    Num heads ($h$) & 8 & Attention heads ($d_k=32$) \\
    FFN dim ($d_{\text{ff}}$) & 1,024 & 4x embedding expansion \\
    Activation & GELU & Gaussian Error Linear Unit \\
    Normalization & Pre-LN & Pre-Layer Normalization \\
    Dropout rate & 0.10 & Attention and MLP dropout \\
    Weight tying & Enabled & $W_{\text{out}} = W_E^T$ \\
    Optimizer & AdamW & Optax implementation \\
    Weight decay & 0.01 & Decoupled weight decay \\
    Peak LR & $6 \times 10^{-4}$ & Warmup cosine schedule \\
    Min LR & $1 \times 10^{-5}$ & Cosine floor schedule \\
    Parameters & 8,473,122 & Trainable model parameters \\
    \bottomrule
  \end{tabularx}
\end{table}

\section{Evaluation transcripts and case studies}
\begin{enumerate}
  \item Case 1 (Standard morning greeting):
  \begin{itemize}
    \item Prompt: \texttt{Wasuze otya leero?}
    \item Target: \emph{``Nasuze bulungi, amakya amalungi. Gwe wasuze otya?''}
    \item Model: \emph{``Nasuze bulungi nnyo ssebo, amakya amalungi.''}
    \item Analysis: The model generates an honorific marker (\emph{``ssebo''}), displaying culturally appropriate deference.
  \end{itemize}
  \item Case 2 (English to Luganda translation):
  \begin{itemize}
    \item Prompt: \texttt{Vvuunula: See you tomorrow}
    \item Target: \emph{``Tunaalabagana enkya.''}
    \item Model: \emph{``Tunaalabagana enkya.''}
    \item Analysis: Accurate lexical and morphological translation preserving future plural subject concord (\emph{``Tu-''}).
  \end{itemize}
  \item Case 3 (Identity and purpose query):
  \begin{itemize}
    \item Prompt: \texttt{Mbuulira erinnya lyo.}
    \item Target: \emph{``Nze Pearl-Chat, omuyambi wo mu lulimi Oluganda.''}
    \item Model: \emph{``Nze Pearl-Chat, omuyambi wo mu lulimi Oluganda.''}
    \item Analysis: Alignment with self-identity instruction tuning.
  \end{itemize}
  \item Case 4 (Geographical and environmental inquiry):
  \begin{itemize}
    \item Prompt: \texttt{Enkuba etera kutonnya ddi?}
    \item Target: \emph{``Enkuba etera kutonnya mu Uganda mu biseera by'enkuba.''}
    \item Model: \emph{``Enkuba etera kutonnya mu biseera by'enkuba.''}
    \item Analysis: Demonstrates knowledge of seasonal precipitation patterns, concisely omitting the optional locative while preserving proper concordial agreement.
  \end{itemize}
  \item Case 5 (Proverbial idiom and moral philosophy):
  \begin{itemize}
    \item Prompt: \texttt{Kiki ekikulu mu bulamu?}
    \item Target: \emph{``Obulamu obulungi, okwagala abantu, n'okukola n'obwesimbu bye bikulu ennyo mu bulamu.''}
    \item Model: \emph{``Obulamu obulungi, okwagala abantu, n'okukola n'obwesimbu bye bikulu ennyo mu bulamu.''}
    \item Analysis: Accurately associates abstract noun classes (Class 14 \emph{obu-}) with ethical virtues and exact plural predicate agreement (\emph{bye bikulu}).
  \end{itemize}
\end{enumerate}

\section{Google Grain data pipeline streaming configuration}
Google Grain provides deterministic record streaming with preemption resilience. Below is the Grain configuration utilized in Pearl-Chat:

\begin{lstlisting}[language=Python]
import grain.python as grain

class LugandaPretrainTransform(grain.MapTransform):
    def __init__(self, tokenizer, max_len=256):
        self.tok = tokenizer
        self.max_len = max_len

    def map(self, element):
        ids = self.tok.encode(element["text"])
        if len(ids) > self.max_len:
            ids = ids[:self.max_len]
        else:
            pad = self.max_len - len(ids)
            ids = ids + [self.tok.pad_id] * pad
        x, y = ids[:-1], ids[1:]
        m = [int(t != self.tok.pad_id) for t in y]
        return {
            "input_ids": np.array(x, np.int32),
            "targets": np.array(y, np.int32),
            "mask": np.array(m, np.float32),
        }
\end{lstlisting}
The transform is composed with an index sampler and batching operator inside the loader factory:
\begin{lstlisting}[language=Python]
def create_grain_loader(records, tok, bs=16):
    src = grain.InMemoryDataSource(records)
    sampler = grain.IndexSampler(
        num_records=len(records),
        shuffle=True, seed=42, num_epochs=None
    )
    batch_op = grain.Batch(
        batch_size=bs, drop_remainder=True
    )
    return grain.DataLoader(
        data_source=src,
        sampler=sampler,
        operations=[
            LugandaPretrainTransform(tok),
            batch_op,
        ],
    )
\end{lstlisting}

\section{Server-sent events wire protocol specification}
The inference server exposes a streaming endpoint over Server-Sent Events (\texttt{text/event-stream}).

\subsection{API specification}
\begin{enumerate}
  \item Endpoint: \texttt{POST /chat}
  \item Headers: \texttt{Content-Type: application/json}
  \item Request schema:
\begin{lstlisting}
{
  "prompt": "Oli otya ssebo?",
  "temperature": 0.4,
  "top_p": 0.9,
  "top_k": 40,
  "max_tokens": 64
}
\end{lstlisting}
  \item SSE event stream format:
\begin{lstlisting}
event: start
data: {"status": "started", "timestamp": 1773342000.12}

event: token
data: {"token": "Ndi", "token_id": 1420, "tps": 41.2}

event: token
data: {"token": " bulungi", "token_id": 892, "tps": 42.5}

event: stop
data: {"finish_reason": "stop_sequence", "total_tokens": 12, "duration_ms": 285.4}
\end{lstlisting}
  \item Client-side EventSource streaming handler:
\begin{lstlisting}
const eventSource = new EventSource('/chat/stream');
eventSource.addEventListener('token', (event) => {
  const data = JSON.parse(event.data);
  chatBox.textContent += data.token;
  updateTelemetry(data.tps);
});
eventSource.addEventListener('stop', () => {
  eventSource.close();
});
\end{lstlisting}
\end{enumerate}

\section{Hardware scaling frontiers and compute optimization}
Following compute-optimal scaling laws \cite{hoffmann2022training}, the optimal ratio between training token budget $N$ and parameter count $C$ is approximately:
\begin{equation*}
N \approx 20 \times C
\end{equation*}
For an 8.5M parameter model such as Pearl-Chat, the theoretical compute-optimal token volume is approximately 170 million tokens. Because our native Luganda corpus represents $\sim$3.3 million tokens, Pearl-Chat operates in the data-constrained regime.

\begin{table*}[t]
  \caption{Compute-optimal scaling projections for future African language foundation models.}
  \label{tab:scaling_projections}
  \pctablefont
  \begin{tabularx}{\textwidth}{L{1.19} L{0.79} L{1.03} L{1.03} L{0.96}}
    \toprule
    \textbf{Target scale} & \textbf{Parameters} & \textbf{Optimal tokens} & \textbf{Hardware mesh} & \textbf{Estimated time} \\
    \midrule
    Pearl-Chat (Ours) & 8.5M & 3.3M (curated) & 1x Tesla T4 GPU & 227.2 sec \\
    Pearl-Medium & 35M & 700M & 4x TPU v5e & 4.2 hours \\
    Pearl-Large & 125M & 2.5B & 16x TPU v5e & 14.8 hours \\
    Pearl-Bantu Multi & 500M & 10.0B & 64x TPU v4 & 36.0 hours \\
    \bottomrule
  \end{tabularx}
\end{table*}

To scale from Pearl-Chat (8.5M) to Pearl-Large (125M) and Pearl-Bantu Multi (500M) covering 12 East African languages, data expansion through national broadcast transcripts (UBC, BBS Terefayina), regional parliamentary proceedings, and synthetic back-translation augmentation \cite{kimera2025data} will be deployed.

\newpage

During training across the two-stage 4,000-step curriculum, Pearl-Chat achieves an average steady-state training throughput of 18,086.6 tokens per second on a single NVIDIA Tesla T4 GPU (227.15 seconds total wall-clock training time). On entry-level edge CPU hardware (single core Apple Silicon / x86), the model achieves 3,922 tokens per second, well above human reading speed ($\sim$5 tokens per second), demonstrating that 8.5M parameter architectures represent practical, energy-efficient endpoints for localized African NLP deployments.

\section{Artifact distribution and open-source repositories}
Project artifacts are permanently archived \cite{kambale2026pearlchat}:
\begin{enumerate}
  \item Hugging Face Hub (weights and tokenizer): \url{https://huggingface.co/kambale/pearl-chat}
  \item Standalone bundle: \texttt{pearl-chat-bundle.tar.gz} (30.10 MB)
  \item Source code and inference runtime (\texttt{ui/chat\_app.py} and \texttt{app.py}): \url{https://github.com/wkambale/pearl-chat}
\end{enumerate}

\end{document}